\documentclass[runningheads]{llncs}

\usepackage{eccv}

\usepackage{eccvabbrv}

\usepackage{graphicx}
\usepackage{booktabs}
\usepackage{arydshln}

\usepackage[accsupp]{axessibility}  

\definecolor{citecolor}{HTML}{2980b9}
\definecolor{linkcolor}{HTML}{c0392b}
\usepackage[pagebackref=true,breaklinks=true,colorlinks,bookmarks=false,citecolor=citecolor,linkcolor=linkcolor]{hyperref}

\usepackage{orcidlink}

\newcommand{\method}{\textbf{ObjectForesight}}

\begin{document}

\title{\textbf{ObjectForesight}: Predicting Future 3D Object Trajectories from Human Videos} 

\titlerunning{ObjectForesight}

\author{Rustin Soraki$^{1}$, Homanga Bharadhwaj$^{2,*}$, Ali Farhadi$^{1,*}$, Roozbeh Mottaghi$^{1,*}$}

\authorrunning{R. Soraki et al.}

\institute{School of Computer Science \& Engineering, University of Washington \and The Robotics Institute, Carnegie Mellon University \\
\email{rustin@cs.washington.edu}}

\maketitle

\begin{center}
    \centering
    \captionsetup{type=figure}
    \includegraphics[width=1\textwidth]{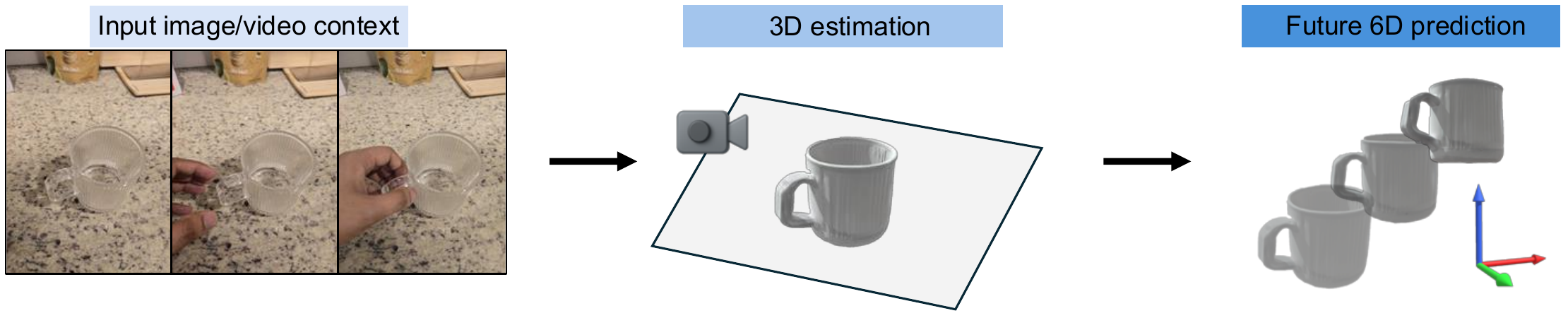}
   \caption{We introduce \method, a framework for predicting future 3D object trajectories from a video context of past motion. We first estimate the object’s 3D shape and initial pose, and then predicts its future 6D poses over time. There are three key contributions: (1) introducing and formalizing the task of 3D object dynamics prediction from human videos, (2) a 3D object-centric dynamics model for future prediction of 6-DoF trajectories, (3) a large-scale dataset of 2+ million object-centric 3D trajectories. Video results are in the website \href{https://objectforesight.github.io/}{objectforesight.github.io}}   
\label{fig:teaser}
\end{center}%

\begin{abstract}
  Humans can effortlessly anticipate how objects might move or change through interaction—imagining a cup being lifted, a knife slicing, or a lid being closed. We aim to endow computational systems with a similar ability to predict plausible future object motions directly from passive visual observation. We introduce \textbf{ObjectForesight}, a 3D object-centric dynamics model that predicts future 6-DoF poses and trajectories of rigid objects from short egocentric video sequences. Unlike conventional world/dynamics models that operate in pixel or latent space, ObjectForesight represents the world explicitly in 3D at the object level, enabling geometrically grounded and temporally coherent predictions that capture object affordances and trajectories. To train such a model at scale, we leverage recent advances in segmentation, mesh reconstruction, and 3D pose estimation to curate a dataset of 2+ million short clips with pseudo-ground-truth 3D object trajectories. Through extensive experiments, we show that ObjectForesight achieves significant gains in accuracy, geometric consistency, and generalization to unseen objects and scenes—establishing a scalable framework for learning physically grounded, object-centric dynamics models directly from observation. \\\href{https://objectforesight.github.io/}{objectforesight.github.io}
  \keywords{3D Future Prediction \and Object-Centric Motion Prediction \and Manipulation Cues from Human Videos}
\end{abstract}

\section{Introduction}
\label{sec:intro}
Humans possess an intuitive understanding of how the world around them can change through interaction. When we see a cup on a table, we can effortlessly imagine it being picked up, tilted, or placed elsewhere. Watching a hand reach toward a knife, we can anticipate the knife’s motion and the transformation of the objects it touches. Such inferences go beyond recognizing what \emph{is} — they reflect our ability to imagine what \emph{can be}. This capacity to mentally simulate object interactions is central to intelligent behavior, allowing us to plan, predict, and act effectively in the physical world \cite{jeannerod2001neural}.

Our goal in this work is to endow computational systems with a similar capability: to infer and predict plausible \emph{future} configurations of objects from passive visual observation. \textbf{We focus on the problem of predicting 3D object dynamics} — \textit{learning how objects can move and interact in 3D space as a result of human actions, without directly modeling the human motion itself}. Rather than learning explicit manipulation trajectories or low-level control policies, we seek to model their \emph{effects}: the diverse, physically coherent object motions that arise from everyday interactions.

To this end, we present~\method, a 3D object-centric forward dynamics model that learns to predict future 6-DoF trajectories of \emph{rigid} objects from egocentric human videos. Given a sequence of RGB frames and an object mesh,~\method~ predicts a temporally coherent sequence of future object poses — effectively imagining how the object may move in the near future (Fig.~\ref{fig:teaser}). Operating in object-centric coordinates allows the model to generalize across varied objects, scenes, and manipulation styles, capturing the underlying semantics of object affordances.

For training~\method, a key challenge is data: There are no large-scale, clean, and physically grounded 3D interaction datasets. Existing robot datasets capture limited, scripted manipulations with explicit action supervision~\cite{rtx}, while internet-scale human video corpora on their own, though rich and diverse, lack aligned 3D information such as object poses, camera geometry, or depth~\cite{smthsmth,Shan20}. To address this, we develop a scalable data curation pipeline that transforms passive human videos into structured 3D motion supervision. Specifically, we extract more than 2 million short clips (2–3 seconds each) from the EPIC-Kitchens dataset~\cite{epic}, automatically detecting hands~\cite{egohos} and identifying objects in contact using SAM~\cite{sam2}. We then recover 3D object meshes and poses with TRELLIS~\cite{trellis2024}, and estimate camera motion and monocular depth using SpaTrackerv2~\cite{spatialtrackerv2}. By expressing object poses relative to the first-frame camera coordinates, we effectively disentangle ego-motion from object motion. This process converts ordinary egocentric videos into a large-scale dataset of 3D object trajectories — the first at this level of scale, fidelity, and semantic diversity.

\method~integrates a Diffusion Transformer (DiT)~\cite{dit} with a geometry aware 3D point encoder, PointTransformerV3~\cite{ptv3}, to jointly reason about object motion and surrounding scene context. Given a short history of RGB frames with corresponding monocular depth maps and a mask of the object in the anchor frame, the model encodes the local 3D geometry of the scene and the object’s recent motion into a unified representation. Conditioned on this visual and spatial context, \method~predicts a distribution over future 6-DoF object poses through a denoising diffusion process. This formulation enables robust, multi-modal prediction of dynamically feasible and physically consistent object motions, maintaining  geometric fidelity and temporal coherence across predicted trajectories.

In summary, \textit{we introduce the task of predicting future 3D object dynamics from videos — a core capability for embodied visual reasoning, and build models and datasets towards this task.} Our key contributions are as follows:
\begin{itemize}
    \item We introduce and formalize the task of \textbf{3D object dynamics prediction from human videos}, establishing a standardized setting for learning how objects move in the real world. This formulation enables models to leverage the vast amount of in-the-wild egocentric video data to learn physical interaction priors without requiring explicit action supervision.
    \item We propose~\method, a 3D object-centric dynamics model that predicts future 6-DoF trajectories of objects from short egocentric video snippets and monocular geometry. 
    \item We construct a large-scale dataset of object-centric 3D trajectories from more than 2 million EPIC-Kitchens clips, using automatic object segmentation and pose estimation to recover high-quality 3D motion supervision from generic interaction videos.
\end{itemize}

Across extensive experiments in daily human activities,~\method~produces accurate, stable, and physically coherent 6-DoF trajectories in diverse real-world scenes. The diffusion-based formulation outperforms autoregressive models and video-generation approaches, offering sharper long-horizon consistency and better multimodal prediction. These results show that large-scale observational data, combined with explicit 3D reasoning, provides a strong foundation for reliable and scalable object-centric motion forecasting. 

\section{Related Works}
\label{sec:related}

\noindent\textbf{Extracting Representations from Human Videos.}
Large-scale egocentric datasets such as Something-Something~\cite{smthsmth}, YouCook~\cite{youcook}, EPIC-Kitchens~\cite{epic}, EGTEA~\cite{egtea}, and Ego4D~\cite{ego4d} have enabled learning rich representations of human–object interactions directly from video. Early work focused on recovering 3D hand and object poses~\cite{zimmermann2017learning, iqbal2018hand, ge20193d, hasson2019learning, rong2020frankmocap} and reconstructing object geometry~\cite{Kehl2017SSD6DMR, Xiang2018PoseCNNAC, He2020PVN3DAD, Hu2019SegmentationDriven6O}, providing geometric supervision for understanding interaction. Advances in tracking and scene flow~\cite{tapir, cotracker, spatialtrackerv2,alltracker} further enable dense motion estimation across time, while recent segmentation and reconstruction systems such as SAM~\cite{sam2}, TRELLIS~\cite{trellis2024}, and very recently SAM3D~\cite{sam3d} make it possible to automatically extract 3D trajectories of objects from in-the-wild images and videos. Our work is closely related in that it leverages these advances to \emph{curate a dataset of object-centric 3D trajectories at scale}, transforming ordinary human videos into a resource for training predictive models of object dynamics. By building upon existing 3D pose estimation and reconstruction pipelines, we focus not on estimating geometry itself, but on learning how objects move and interact over time.

\noindent\textbf{Predicting Manipulation Cues from Human Videos.}
Another line of research focuses on predicting or reasoning about manipulation cues and affordances from human videos. Classical works in affordance learning~\cite{nagarajan2019grounded, brahmbhatt2019contactgrasp, mo2021where2act, liu2022joint, handsasprobes,Mottaghi2015NewtonianIU} study how objects are grasped, where contact occurs, how hands move in the future~\cite{handsonvlm,egoman,liu2022joint} or which parts of an object afford specific actions. More recent approaches~\cite{Shan20, cmudata, track2act} learn to anticipate manipulation outcomes or future contact regions, connecting perception to physical reasoning. Such methods primarily operate in 2D or intermediate feature space, forecasting human or object-centric cues that signal future interactions. Our work shares the goal of extracting predictive signals from human videos but differs in focus. Rather than predicting contact maps or categorical actions, we aim to learn a continuous model of \emph{3D object dynamics}—how objects themselves move in space as a result of human interactions. By grounding prediction in SE(3) pose space and explicit geometry, we extend affordance learning toward physically coherent, object-centric reasoning about future motion.

\noindent\textbf{World Models and Trajectory Representations.}
Building models of how the world evolves in response to interaction has long been a core challenge in both computer vision and robotics. Recent efforts in visual world modeling have primarily focused on learning predictive representations either at the pixel level through video generation~\cite{villegas2017decomposing, walker2016uncertain} or in latent spaces through representation learning~\cite{Hafner2019Dreamer, Kipf2019Contrastive,vjepa2}. While such approaches capture temporal dependencies, they often lack explicit 3D grounding and object-level motion prediction. In contrast, our work develops an \textit{explicit 3D object trajectory model} that operates in SE(3) space. Instead of predicting future pixels or abstract latent codes, our method explicitly models object evolution in 6-DoF pose space, and unlike implicit language conditioning~\cite{egoscaler}, conditions explicitly on predicted object geometry and past motion context, offering a physically grounded representation well-suited for integration into robotic manipulation frameworks~\cite{maniptrans,track2act}. Future prediction has been explored in the self-driving scenarios (e.g., \cite{gu2023vip3d}), but they focus on largely independent agents and external objects in a different dynamics regime.

\section{Method}
\label{sec:method}

\begin{figure*}[t]
    \centering
    \includegraphics[width=\textwidth]{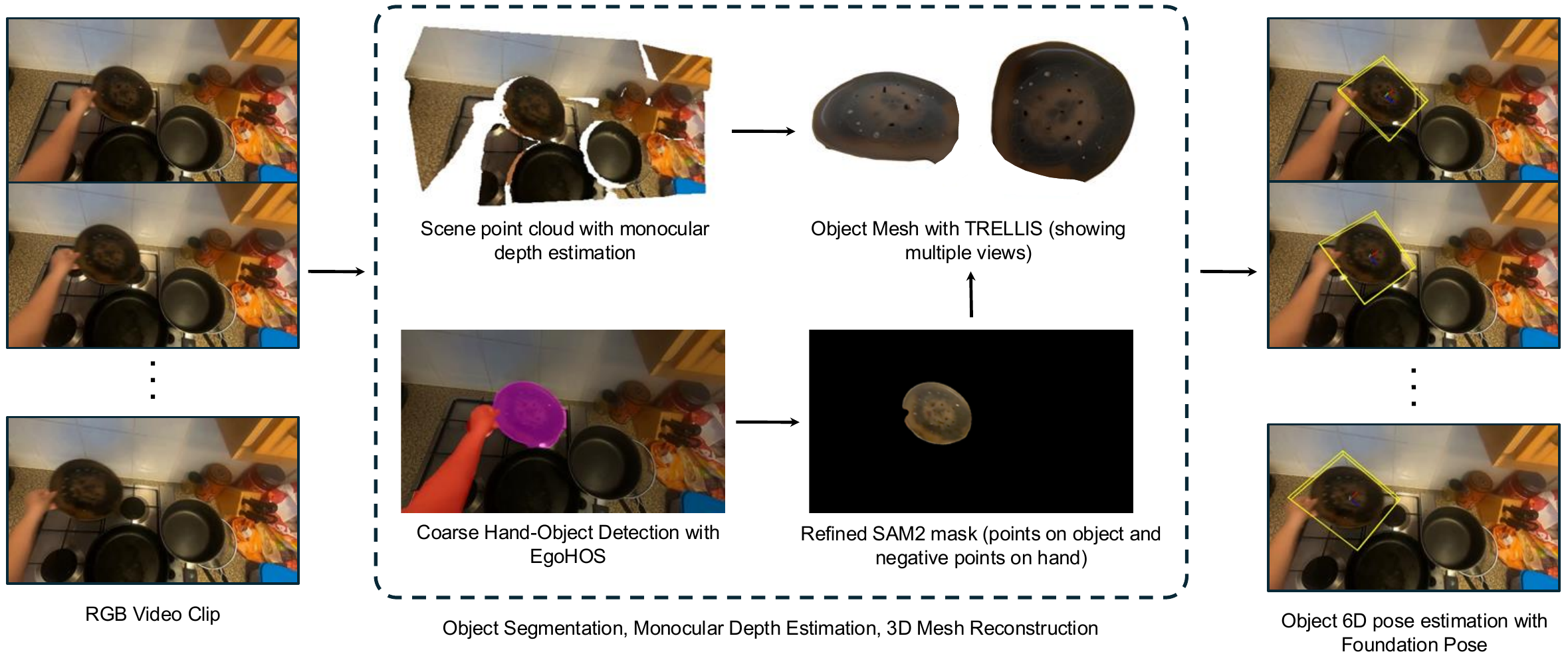}
    \vspace{-0.7cm}
    \caption{\textbf{Data curation pipeline from egocentric video to 3D object trajectories.}
Starting from EPIC-Kitchens action segments, we detect hands and objects, refine masks, and filter for clear manipulations.  
We then reconstruct an object mesh, recover metric depth and camera geometry, and do 6-DoF pose estimation and tracking.  
Sliding windows over these tracks yield short, clean, anchor-frame–canonicalized 6-DoF trajectories used to train \method.}
    \vspace{-0.4cm}
    \label{fig:datacuration}
\end{figure*}

We aim to learn a forward dynamics model that predicts future 3D poses of rigid objects from passive human videos. The task involves inferring plausible 6-DoF trajectories conditioned on observed object geometry, local scene context, and a short history of object motion. Since no dataset exists for this setting, we construct a large-scale dataset of 3D object trajectories from egocentric human activity videos using off-the-shelf vision models (Sec.~\ref{sec:data}). We then train a diffusion-based transformer model (Sec.~\ref{sec:forecast}) that learns to sample diverse, physically consistent future trajectories conditioned on visual and geometric context.

\subsection{Overview}
\label{sec:overview}

\method\ tackles the problem of predicting future 3D object motion from short windows of egocentric video.  
Given $C$ observed frames and a prediction horizon of $H$, the goal is to model a distribution over the next $H$ future 6-DoF poses of a manipulated object.  
All frames in the window are expressed in the anchor-frame (first frame of the prediction horizon) camera coordinates, allowing us to isolate true object motion from ego-motion.
In our default setting, we use $C{=}3$ and $H{=}8$.

Formally, we observe images $\mathcal{I}_{1:C}$ and their corresponding object poses
\[
\mathbf{P}_{1:C}=[\mathbf{p}_1,\dots,\mathbf{p}_C], \qquad \mathbf{p}_t \in \mathrm{SE(3)},
\]
where each pose token
$\mathbf{p}_t = [x_t,y_t,z_t,\mathbf{r}_{t,\mathrm{6D}}]$
contains translation and a continuous 6D rotation representation~\cite{Zhou2018OnTC}.  
Depth from the anchor frame is backprojected to form a point cloud $\mathbf{X}$, and normalized object bounding boxes $\mathbf{B}_{1:P}$ provide coarse spatial cues.  
The forecasting target is the future sequence
\[
\mathbf{P}_{\text{future}} = [\mathbf{p}_{t_a},\dots,\mathbf{p}_{t_a+H-1}],
\quad t_a=C{+}1.
\]

\method~ contributes both \textit{data} and \textit{modeling}:  
(i) a large-scale pipeline that converts raw egocentric videos into metrically grounded, anchor-frame–canonicalized 6-DoF trajectories; and  
(ii) a geometry-aware diffusion model that predicts future object motion from these trajectories.

Our predictive architecture combines a context-conditioned geometry encoder over the anchor-frame point cloud with a Diffusion Transformer (DiT) temporal backbone. The encoder conditions point features on the recent motion context (FiLM) and pools them into an object-centric scene embedding $\mathbf{z}_{\mathrm{geom}}$, while the DiT models a distribution over future pose sequences conditioned on $\mathbf{z}_{\mathrm{geom}}$ and an explicit pose-token prefix.

The model operates in a depth-normalized pose space for stability, and uses a cosine noise schedule with v-parameterized denoising.  
At inference, DDIM sampling produces smooth, diverse, and physically coherent 3D trajectories.

\subsection{Data Curation: From Egocentric Video to 3D 6-DoF Object Trajectories}
\label{sec:data}

Our curation pipeline converts in-the-wild egocentric videos into clean,
metrically grounded trajectories of hand-manipulated objects
(Fig.~\ref{fig:datacuration}).  Starting from EPIC-Kitchens action segments, we
apply a sequence of automatic extraction and quality gates to recover
temporally coherent 6-DoF poses.  We summarize the key stages below.\\

\noindent\textbf{Action segment prefiltering.} 
We begin from annotated single-activity segments and discard clips longer than 10~seconds to limit drift and ensure short, interaction-centric windows. \\

\noindent\textbf{Hand--object discovery with EgoHOS.}
For each remaining clip, we run EgoHOS~\cite{egohos} to segment hands and
candidate manipulated objects frame-wise.  Frames without hands or without any
object hypotheses are removed.  This yields per-frame masks for (i) active
hand(s) and (ii) plausible manipulated objects.\\

\noindent\textbf{Robust object masks with temporal consensus.}
We initialize SAM2~\cite{sam2} using point prompts derived from EgoHOS masks
and propagate a single object instance through the clip.  Positive prompts
come from the interior of the EgoHOS object mask; negative prompts are drawn
from the hand mask, the other-hand object (if present), and a thin ring around
the object boundary.  To mitigate occasional EgoHOS failures, we form
\emph{temporal consensus prompts}, intersections of masks over a small temporal
window, which bias SAM2 toward temporally stable shapes.  Newly proposed SAM2
masks must have low IoU with the active tracks to prevent duplication.  The
result is a temporally smooth, occlusion-resilient object mask sequence.\\

\noindent\textbf{VLM gating for manipulation and view quality.}
We apply a two-stage VLM-based filter using InternVL3~\cite{internvl3}.  
First, at the video level, we check whether the highlighted object is actually
moved by hand; static objects are discarded.  
Second, at the frame level, we crop around the object and evaluate visibility
(no blur, limited occlusion).  
Frames passing this test form the set of \emph{clean views}.\\

\noindent\textbf{Object 3D reconstruction from clean views.}
TRELLIS~\cite{trellis2024} reconstructs a 3D object mesh from clean views.
The mesh is \emph{not} used during \method\ training; it only serves as a
geometric template for model-based pose estimation under occlusion.\\

\noindent\textbf{Model-based 6-DoF pose with metric depth and amodal masks.}
SpaTrackerV2~\cite{spatialtrackerv2} provides metric depth and camera geometry.
We use Diffusion-VAS~\cite{vas} to complete amodal object masks.  
Pose initialization and tracking use FoundationPose~\cite{Wen2023FoundationPoseU6}
with three modifications for egocentric video:

(i)~\emph{Metric scale estimation.}
TRELLIS meshes lack scale; we estimate scale by comparing masked depth points
to mesh radii across neighboring frames (robust weighted median), then refine
via depth–silhouette alignment.

(ii)~\emph{Multi-view initialization.}
We pick up to five clean views, run FoundationPose initialization, and refine
each using depth alignment and silhouette consistency.  We choose the best by
FoundationPose score, with an IoU-based override. Low-IoU cases are
discarded.

(iii)~\emph{Bidirectional tracking with re-registration.}
From the best initialization we track forward and backward.  
If projection IoU drops below~0.1, we trigger local re-registration using the
current mask.  
This produces temporally coherent pose tracks with explicit re-registration
events.\\

\noindent\textbf{Trajectory slicing and final quality control.}
We slide a window of length $C{+}H$ along each track.  
A window is kept if it lies within a single registration segment and maintains
stable projection IoU (no drop $>0.1$).  
All poses are re-expressed in the anchor-frame camera coordinates to remove
ego-motion.\\

\noindent\textbf{Outcome.}
This automatic pipeline enforces (i)~manipulation validity (VLM gating),
(ii)~mask fidelity (SAM2 with temporal consensus and amodal completion), and
(iii)~metric, temporally coherent poses (FoundationPose with depth, geometry,
and re-registration).  
The result is a large collection of short, clean, object-centric trajectories
suitable for training multi-modal 3D dynamics models.

\begin{figure*}[tp]
    \centering
    \includegraphics[width=\textwidth]{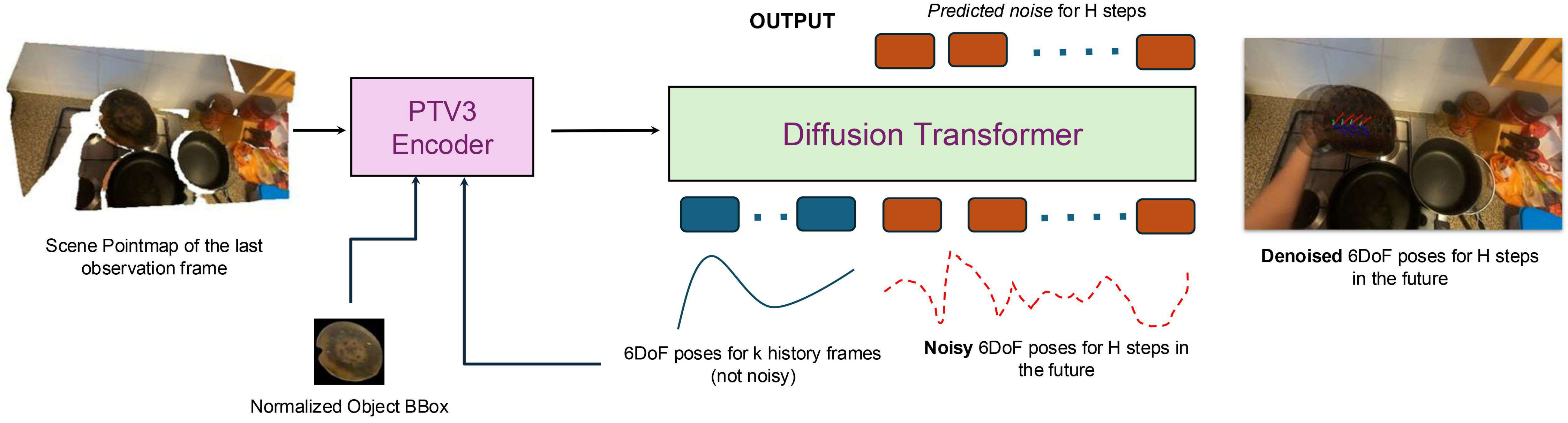}
   \vspace{-0.5cm}
   \caption{\textbf{Model architecture.}  
        Given past pose tokens and their normalized bounding boxes, we summarize motion context with anchor-query attention and use it to guide object-centric pooling in a PointTransformerV3 encoder, producing a geometry-aware scene embedding. A diffusion transformer (DiT, AdaLN-Zero) then denoises future depth-normalized pose tokens, conditioned on the scene embedding and an explicit prefix of past pose tokens. This design allows \method\ to generate diverse, physically coherent, and temporally smooth 3D motion predictions.}
    \vspace{-0.3cm}
    \label{fig:architecture}
\end{figure*}

\subsection{Predicting Future Trajectories in 3D}
\label{sec:forecast}

Our forecasting model learns to generate diverse and physically coherent future
pose sequences conditioned on the current geometric and motion context.  It
combines a geometry-aware encoder with a diffusion-based transformer operating
on object-centric, depth-normalized 9D pose tokens. An outline of the model is depicted in Fig.~\ref{fig:architecture}. \\

\noindent\textbf{Scene and Context Encoding.}
Given an anchor-frame point cloud $\mathbf{X} \in \mathbb{R}^{N \times 3}$,
conditioning pose tokens $\mathbf{P}_{1:t_a} \in \mathbb{R}^{t_a \times 9}$,
and corresponding normalized bounding boxes $\mathbf{B}_{1:t_a} \in \mathbb{R}^{t_a \times 4}$, our goal is to construct a compact representation that summarizes both the
recent motion and the 3D scene structure. Here $N$ is the number of points sampled from the anchor-frame depth map. We use $C$ pre-anchor context frames, so the anchor index is $t_a=C{+}1$ in the $C+H$ window.

For each frame $k$ in the conditioning sequence, we form $[P_k B_k] \in \mathbb{R}^{13}$ and project it into a $64$D context space. We then pool the conditioning sequence with attention: the anchor token queries all
conditioning tokens, and we add a sinusoidal embedding of the relative time to
the anchor with a learnable scale.  This yields a single context vector $\mathbf{ctx} \in \mathbb{R}^{64}$.

We feed the point cloud $\mathbf{X}$ into a PointTransformerV3 encoder~\cite{ptv3}. Each point is represented by its anchor-camera coordinates and its coordinates in the estimated anchor object frame, enabling object-centric reasoning. We also provide the encoder with $\mathbf{ctx}$ which conditions the point cloud features on it using feature-wise linear modulation (FiLM)~\cite{perez2018film}. We then pool point features into a global scene embedding $\mathbf{z}_{\mathrm{geom}} \in \mathbb{R}^{512}$ using an object-centric attention head that matches point features to a query derived from $\mathbf{ctx}$ and biases weights toward points near the object. $\mathbf{z}_{\mathrm{geom}}$ is then used as a conditioning signal, injected using AdaLN-Zero~\cite{dit} inside the DiT  blocks. \\

\noindent\textbf{Tokenization of Pose Sequences.}
We operate on object-centric pose tokens expressed in the anchor-frame camera
coordinates.  Each pose $\mathbf{p}_t = [x_t, y_t, z_t, \mathbf{r}_{t,\mathrm{6D}}]$
is reparameterized into a depth-normalized token:
\[
    \mathbf{y}_t
    =
    [u_t, v_t, s_t, \mathbf{r}_{t,\mathrm{6D}}],
    \quad
    u_t=\tfrac{x_t}{z_t},\;
    v_t=\tfrac{y_t}{z_t},\;
    s_t=\log z_t,
\]
which reduces the dynamic range of translation and improves numerical
stability.  For the future horizon of length $H$, we form
$\mathbf{Y}_{\mathrm{future}} = [\mathbf{y}_{t_a},\dots,\mathbf{y}_{t_a+H-1}]$.
We then apply channel-wise standardization using statistics
$(\boldsymbol{\mu}, \boldsymbol{\sigma})\in\mathbb{R}^9$ estimated over the first training batches (and fixed thereafter).

We apply the same depth reparameterization and standardization to the
context-frame poses, yielding normalized context tokens $\tilde{\mathbf{P}}_{1:t_a}$. These tokens are embedded and prepended as a prefix to the future sequence inside the transformer, giving the DiT access to the full conditioning history. \\

\noindent\textbf{Forward Diffusion Process and Cosine Schedule.}
Let $\tilde{\mathbf{Y}}_0 \in \mathbb{R}^{H \times 9}$ be the clean
normalized future sequence for the batch. Following the standard diffusion
framework, we define a forward noising process with a cosine $\beta$-schedule and
sample diffusion timesteps $\tau$ uniformly from $\{0,\dots,T{-}1\}$ (we use $T{=}1000$).
The DiT processes the noised sequence $\tilde{\mathbf{Y}}_{\tau}$ as a length-$H$
sequence of 9D tokens, conditioned on the diffusion-timestep embedding, the
geometric embedding $\mathbf{z}_{\mathrm{geom}}$, and the normalized context tokens
$\tilde{\mathbf{P}}_{1:t_a}$.

Tokens are embedded into a latent sequence and augmented with learned absolute positions, a token-type embedding (context vs.\ future), and a signed anchor-relative time embedding. Conditioning is injected via AdaLN-Zero where a lightweight MLP combines timestep and scene embeddings into per-layer normalization modulations and gated residuals within each transformer block. \\

\noindent\textbf{v-Parameterization with p2 Weighting.}
Instead of predicting the noise $\boldsymbol{\epsilon}$ directly, we adopt
v-parameterization, which stabilizes training across timesteps. We train with an SNR-weighted regression loss (p2 reweighting~\cite{stablediffusion}) and additionally apply horizon-aware weighting that linearly increases toward later forecast steps (from 1 to 3 across the horizon). During training and DDIM sampling, we reconstruct $\hat{\tilde{\mathbf{Y}}}_0$ from the predicted $\mathbf{v}_\theta$ using the standard closed-form relation. \\

\noindent\textbf{De-normalization and Pose Decoding.}
We recover physical 9D poses by reversing the standardization and depth
reparameterization applied during preprocessing. The network output is
first de-standardized as
\begin{equation}
    \hat{\mathbf{Y}}_0
    =
    \hat{\tilde{\mathbf{Y}}}_0 \odot \boldsymbol{\sigma}
    +
    \boldsymbol{\mu},
\end{equation}
where $\odot$ is elementwise multiplication. Each resulting token
$\hat{\mathbf{y}}_t = [\hat{u}_t,\hat{v}_t,\hat{s}_t,
\hat{\mathbf{r}}_{t,\mathrm{6D}}]$ is then converted back to Cartesian
translation coordinates by inverting the log-depth and
normalized-coordinate transforms:
\[
    \hat{z}_t = \exp(\hat{s}_t),
    \qquad
    \hat{x}_t = \hat{u}_t\,\hat{z}_t,
    \qquad
    \hat{y}_t = \hat{v}_t\,\hat{z}_t.
\]
This yields the final 9D pose token
$[\hat{x}_t,\hat{y}_t,\hat{z}_t,\hat{\mathbf{r}}_{t,\mathrm{6D}}]$. \\

\noindent\textbf{Losses.}
The primary training objective is the v-prediction MSE,
\[
    \mathcal{L}_{v}
    =
    \mathbb{E}\!\left[
        w(\tau)\,\left\|\mathbf{v}_\theta(\tilde{\mathbf{Y}}_\tau, \tau)
        - \mathbf{v}_\tau\right\|_2^2
    \right],
\]
where
$\mathbf{v}_\tau
= \sqrt{\bar{\alpha}_\tau}\,\boldsymbol{\epsilon}
- \sqrt{1-\bar{\alpha}_\tau}\,\mathbf{Y}_0$
is the v-parameterization target and
$w(\tau) = (1 + \mathrm{SNR}(\tau))^{-\gamma}$ is a P2-style weight that
downweights low-noise timesteps.

Because we predict poses in anchor-frame camera coordinates, we can
additionally supervise the decoded SE(3) trajectory directly. For each
future step $k$, we convert the predicted 6D rotation to
$\hat{\mathbf{R}}_k \in \mathrm{SO}(3)$ and measure translation error
$\|\mathbf{t}_k - \hat{\mathbf{t}}_k\|_2$ and rotation error via the
geodesic distance
$\mathrm{d}_{\mathrm{geo}}(\mathbf{R}_k, \hat{\mathbf{R}}_k)$.
Averaging both over the prediction horizon gives the auxiliary pose loss
\[
    \mathcal{L}_{\mathrm{aux}}
    =
    \mathbb{E}\!\left[
        \bar{\alpha}_\tau\!\left(
            \lambda_R\,\overline{d}_{\mathrm{geo}}
            + \lambda_{\mathrm{trans}}\,\overline{e}_{\mathrm{trans}}
        \right)
    \right],
\]
where $\overline{d}_{\mathrm{geo}}$ and
$\overline{e}_{\mathrm{trans}}$ are the horizon-averaged rotation
(radians) and translation errors, respectively. The expectation is over
training samples and sampled diffusion steps;
$\lambda_R$ and $\lambda_{\mathrm{trans}}$ balance the two error
magnitudes; and the factor $\bar{\alpha}_\tau$ (near $1$ for clean
samples, near $0$ for noisy ones) downweights steps where the
$\hat{\tilde{\mathbf{Y}}}_0$ reconstruction is unreliable.

To encourage smooth trajectories we add SE(3) velocity and acceleration
losses on consecutive pose increments (also weighted by
$\bar{\alpha}_\tau$). Defining translation increments
$\Delta\mathbf{t}_k = \mathbf{t}_{k+1} - \mathbf{t}_k$, rotation
increments $\Delta\mathbf{R}_k = \mathbf{R}_k^\top\mathbf{R}_{k+1}$,
and letting $\Delta^2$ denote second differences:
\[
    \begin{aligned}
        \mathcal{L}_{\mathrm{vel}}
        &=
        \overline{
            \|\Delta\mathbf{t}_k - \Delta\hat{\mathbf{t}}_k\|_2^2
            + \mathrm{d}_{\mathrm{geo}}(
                \Delta\mathbf{R}_k, \Delta\hat{\mathbf{R}}_k)^2
        },\\[4pt]
        \mathcal{L}_{\mathrm{acc}}
        &=
        \overline{
            \|\Delta^2\mathbf{t}_k - \Delta^2\hat{\mathbf{t}}_k\|_2^2
            + \mathrm{d}_{\mathrm{geo}}(
                \Delta^2\mathbf{R}_k, \Delta^2\hat{\mathbf{R}}_k)^2
        },
    \end{aligned}
\]
where $\overline{\,\cdot\,}$ averages over valid horizon steps $k$.
Finally, a small depth-floor penalty discourages degenerate predictions
with extremely small depth:
\[
    \mathcal{L}_{z_{\min}}
    =
    0.01\,\mathrm{ReLU}(z_{\min} - \hat{z}_t).
\]

The complete objective is
\[
    \mathcal{L}_{\mathrm{total}}
    =
    \mathcal{L}_{\mathrm{v}}
    + \mathcal{L}_{\mathrm{aux}}
    + \mathcal{L}_{z_{\min}}
    + \lambda_{\mathrm{vel}}\,\mathcal{L}_{\mathrm{vel}}
    + \lambda_{\mathrm{acc}}\,\mathcal{L}_{\mathrm{acc}},
\]
where $\mathcal{L}_{\mathrm{v}}$ operates in normalized token space
while all other terms are computed on the decoded pose sequence (after
de-normalization and depth decoding). We set $\lambda_R = 2.0$,
$\lambda_{\mathrm{trans}} = 20.0$, $\lambda_{\mathrm{vel}} = 0.5$, and
$\lambda_{\mathrm{acc}} = 0.1$. \\

\noindent\textbf{Diffusion Sampling}
At inference time we sample from pure Gaussian noise using deterministic DDIM with $S{=}50$ evenly spaced denoising steps. At each step the model predicts $\mathbf{v}_\theta$ conditioned on $(\mathbf{z}_{\mathrm{geom}}, \tilde{\mathbf{P}}_{1:t_a})$, from which we reconstruct $\hat{\tilde{\mathbf{Y}}}_0$ and apply the DDIM update. After the final step, the result is the predicted future trajectory $\hat{\mathbf{P}}_{\mathrm{future}}$ in the anchor frame. \\

\noindent\textbf{Why Diffusion?}
Diffusion-based modeling is well suited to 3D interaction dynamics.  Given
identical 3D conditioning, multiple future motions can be plausible (e.g., a
mug can be picked up, slid, or rotated).  Our DiT captures this inherently
one-to-many nature while encouraging temporally smooth, physically plausible
trajectories. In our experiments, it yields more plausible and geometrically
consistent predictions than an autoregressive transformer baseline trained on
the same object-centric representation. \\

\noindent\textbf{Summary.}
By combining an object-centric 3D scene encoder with an AdaLN-Zero conditioned
diffusion transformer over depth-normalized pose tokens, \method\ learns a
rich conditional distribution over future object motion.  The architecture
explicitly leverages metric geometry, camera coordinates, and pose history to
generate accurate, diverse, and physically plausible 6-DoF trajectories in
real-world egocentric scenes.

\section{Experiments}
\label{sec:experiment}

Our experiments aim to answer three questions:  
\begin{enumerate}
    \item Is the curated dataset of object-centric 3D trajectories reliable and diverse? 
    \item Are the predicted future object motions plausible and physically consistent? 
    \item Does the model generalize beyond the distribution of curated scenes?
\end{enumerate}

\subsection{Datasets}
\begin{figure*}[t]
    \centering
    \begin{subfigure}[c]{0.50\textwidth}
        \centering
        \includegraphics[width=\textwidth]{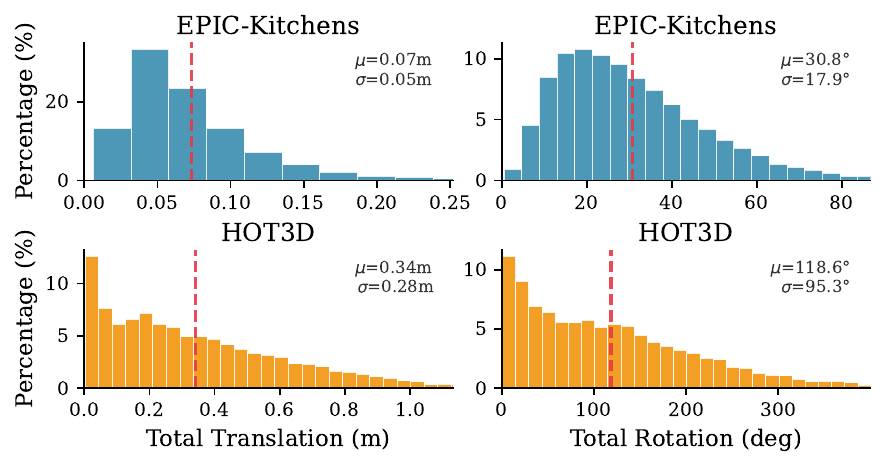}
        \caption{}
        \label{fig:motion_stats}
    \end{subfigure}
    \hfill
    \begin{subfigure}[c]{0.48\textwidth}
        \centering
        \scriptsize
        \setlength{\tabcolsep}{2pt}
        \begin{tabular}{l r}
        \toprule
        \textbf{Step Name} & \textbf{Number} \\
        \midrule
        Action Segments & 76,885 vids \\
        Selected Vids (hands, $\leq10$s) & 72,046 vids \\
        SAM2 Tracks & 229,102 tracks \\
        Filtered Tracks & 112,057 tracks \\
        TRELLIS Models & 71,296 models \\
        Objects with Pose Tracks & 59,174 tracks \\
        Pre-Filter Trajectories & 3,065,568 trajs \\
        Post-Filter Trajectories & 2,073,109 trajs \\
        \bottomrule
        \end{tabular}
        \caption{}
        \label{fig:curation_stats}
    \end{subfigure}
    \vspace{-0.3cm}
    \caption{\textbf{Dataset overview.} (a) Motion statistics of the curated 3D object trajectories and HOT3D dataset, showing total translation and rotation for all objects.  (b) Summary of each stage in our automatic data curation pipeline as applied to EpicKitchens videos.}
    \label{fig:dataset_stats}
\end{figure*}

We use our curated dataset from EpicKitchen for the main experiments. We also conducted further experiments with the HOT3D-Clips~\cite{banerjee2024hot3d} dataset that includes larger object motion and precise groundtruth but collected in a lab setting. Motion statistics, along with the mean and standard deviation, of the object translation and rotation in the trajectories from both datasets can be seen in Fig.~\ref{fig:motion_stats}. \\

\noindent\textbf{Curated EpicKitchen Dataset Details.} We curate a large-scale collection of object-centric 3D motion trajectories from egocentric videos using an automated eight-stage pipeline (Fig.~\ref{fig:curation_stats}). From \textbf{76K} EPIC-Kitchens action segments, we retain \textbf{72K} short clips ($\leq$10s) with visible hands to ensure the presence of interactions. Object masks and 2D tracks are obtained using SAM2, yielding \textbf{229K} raw tracks before quality filtering reduces them to \textbf{112K}. Using TRELLIS, we reconstruct \textbf{71K} object meshes and obtain \textbf{59K} pose-aligned tracks. Sliding-window extraction produces \textbf{3.06M} raw (3+8)-step 3D trajectories, which are further filtered to \textbf{2.07M} high-quality trajectories, each consisting of a 0.13s window, used for training and evaluation. \\

\noindent\textbf{HOT3D-Clips.} We also train and evaluate \method\ on HOT3D-Clips to validate that the model can learn future 3D motion from cleaner trajectories. For the HOT3D experiments, we skip frames to convert the clips to 6 fps and then extract the same $(3{+}8)$-frame windows used in our main setting, making the trajectories 1.33s long. We also filter-out trajectories where the object is either stationary or the movement is negligible (less 0.01m translation or less than 5\textdegree\ rotation). This gives us \textbf{167K} high-quality trajectories with large object state change.


\subsection{Baselines, Ablations, and Metrics}

We compare four approaches.
\textbf{ObjectForesight-DiT} is our diffusion transformer for
multimodal trajectory prediction.
\textbf{ObjectForesight-AR} is an autoregressive transformer variant
that removes the diffusion formulation.
\textbf{Constant Velocity} is a simple baseline that extrapolates
future translation and rotation assuming the velocity observed in the
context frames remains constant.
Finally, \textbf{Video-generation (Luma Ray3)} is an off-the-shelf state-of-the-art
video generator that synthesizes a short future clip from three context
frames, from which we recover 6-DoF poses using our curation pipeline.
Because this pipeline is computationally expensive, we evaluate it on
20 randomly selected validation videos with clear object visibility.

We report six trajectory-level metrics. For translation:
\textbf{ADE} (average displacement error across all timesteps),
\textbf{FDE} (displacement error at the final timestep), and
\textbf{DES} (displacement error slope, capturing the per-timestep
trend); all three are in meters. For rotation:
\textbf{ARE} (average rotation error),
\textbf{FRE} (final rotation error), and
\textbf{RES} (rotation error slope); all three are in degrees.
Additional ablations, such as varying the number of history frames,
are provided in the supplementary material.

\begin{figure*}[t]
    \centering
    \includegraphics[width=\textwidth]{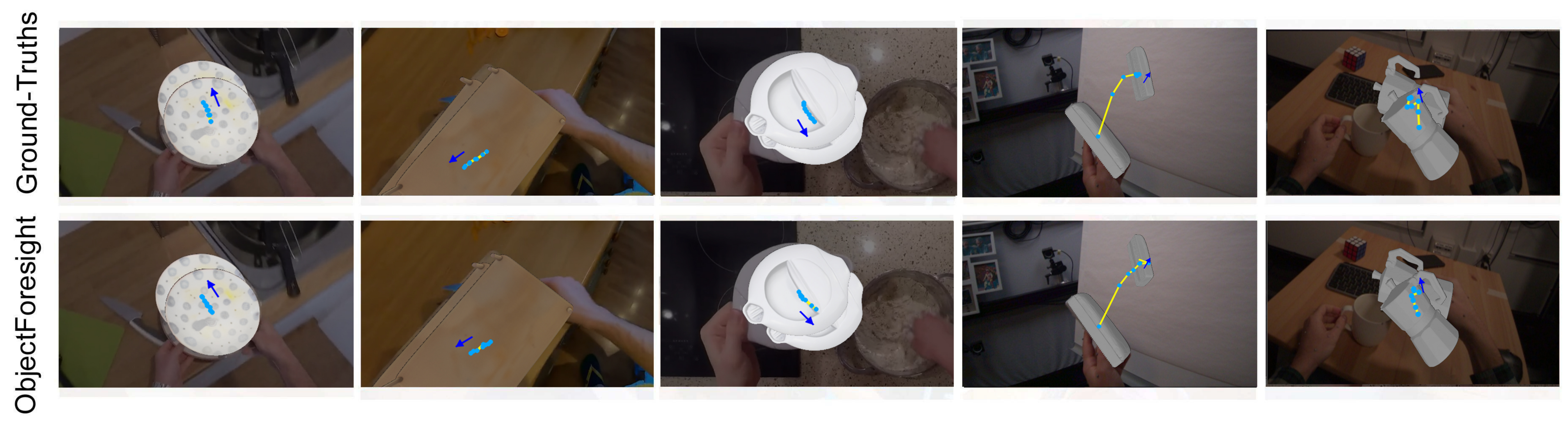}
   \caption{\textbf{Qualitative results from \method.}  
From the left, the first three columns are from the curated dataset from EpicKitchen and the final two columns are from the HOT3D-Clips dataset. For each sequence, we overlay 8 predicted poses on the last observed frame, illustrating the projected object centers, the transformed object mesh, and the general direction of the movement using an arrow. 
The images are zoomed in for clarity. }
    \label{fig:qualitative}
\end{figure*}



\subsection{Qualitative Results}
Fig.~\ref{fig:qualitative} shows that ObjectForesight produces smooth,
physically plausible 6-DoF trajectories across diverse manipulation
scenarios, capturing realistic interactions such as lifting, rotating,
and placing objects while maintaining temporal coherence with the
observed context. The examples span a range of motions, from simple translations of a plate, to the more complex rotation of a kettle, and the wiping action of the eraser on the board. For videos, refer to the supplementary material. 

\setlength{\tabcolsep}{6pt}
\begin{table*}[t]
\centering
\scriptsize
\begin{tabular}{lcccccc}
\toprule
 & ADE\,$\downarrow$ & FDE\,$\downarrow$ & DES$\,\downarrow$ & ARE$\,\downarrow$ & FRE$\,\downarrow$ & RES$\,\downarrow$ \\
\midrule
\multicolumn{7}{l}{\textit{Epic-Kitchens}} \\
\addlinespace[2pt]
Constant Velocity & 0.027 & 0.053 & 0.007 & 2.47\textdegree & 5.60\textdegree & 0.80\textdegree \\
ObjectForesight-AR & 0.067 & 0.074 & \textbf{0.002} & 9.48\textdegree & 12.58\textdegree & 0.93\textdegree \\
\textbf{ObjectForesight-DiT} & \textbf{0.016} & \textbf{0.029} & 0.004 & \textbf{2.30\textdegree} & \textbf{4.82\textdegree} & \textbf{0.66\textdegree} \\
\addlinespace[2pt]
\hdashline
\addlinespace[2pt]
\quad \textit{vs.\ Video Generation} &  &  & \\
\addlinespace[2pt]
Luma AI Ray3 & 0.084 & 0.149 & 0.020 & 12.86\textdegree & 20.90\textdegree & 2.62\textdegree \\
\textbf{ObjectForesight-DiT} & \textbf{0.029} & \textbf{0.059} & \textbf{0.008} & \textbf{7.29\textdegree} & \textbf{13.98\textdegree} & \textbf{1.77\textdegree} \\
\midrule
\multicolumn{7}{l}{\textit{HOT3D-Clips}} \\
\addlinespace[2pt]
Constant Velocity & 0.136 & 0.280 & 0.04 & 38.70\textdegree & 68.53\textdegree & 9.85\textdegree \\
ObjectForesight-AR & 0.055 & 0.082 & 0.007 & 9.80\textdegree & 14.95\textdegree & 1.55\textdegree \\
\textbf{ObjectForesight-DiT} & \textbf{0.021} & \textbf{0.026} & \textbf{0.003} & \textbf{8.92\textdegree} & \textbf{12.58\textdegree} & \textbf{1.16\textdegree} \\
\bottomrule
\end{tabular}
\vspace{0.2cm}
\caption{
\textbf{3D trajectory forecasting on Epic-Kitchens and
HOT3D-Clips.} Lower is better for all metrics.
ObjectForesight-DiT outperforms or matches ObjectForesight-AR across all
metrics on both datasets, and substantially outperforms the Luma AI Ray3
video-generation baseline on Epic-Kitchens, highlighting the benefit of
modeling motion directly in SE(3) rather than inferring it from
synthesized frames. Note that the comparison with the video generation method is conducted on a subset of the dataset due to its computational complexity.}
\label{tab:quantitative}
\end{table*}

\setlength{\tabcolsep}{1pt}
\begin{table*}[t]
\centering
\scriptsize
\begin{subtable}[t]{0.49\textwidth}
\centering
\caption{Scene Encoders}
\vspace{-0.15cm}
\begin{tabular}{lcccc}
\toprule
 & ADE\,$\downarrow$ & FDE\,$\downarrow$ & ARE$\,\downarrow$ & FRE$\,\downarrow$ \\
\midrule
DGCNN~\cite{wang2019dynamicgraphcnnlearning} & 0.0171 & 0.0297 & 2.333\textdegree & 4.968\textdegree \\
PointNet++~\cite{qi2017pointnetdeephierarchicalfeature} & 0.0171 & 0.0298 & 2.357\textdegree & 5.024\textdegree \\
SparseConv & 0.0179 & 0.0295 & 2.700\textdegree & 5.299\textdegree \\
No-Encoder & 0.0174 & 0.0298 & 2.690\textdegree & 5.380\textdegree \\
\addlinespace[2pt]
\hdashline
\addlinespace[2pt]
\textbf{PTV3} & \textbf{0.0165} & \textbf{0.0287} & \textbf{2.299\textdegree} & \textbf{4.816\textdegree} \\
\bottomrule
\end{tabular}
\label{tab:ablation_encoder}
\end{subtable}
\hfill
\begin{subtable}[t]{0.49\textwidth}
\centering
\caption{DiT Scaling}
\vspace{-0.15cm}
\begin{tabular}{lcccc}
\toprule
 & ADE\,$\downarrow$ & FDE\,$\downarrow$ & ARE$\,\downarrow$ & FRE$\,\downarrow$ \\
\midrule
6L-384D & 0.0193 & 0.0311 & 4.242\textdegree & 7.762\textdegree \\
8L-512D & 0.0171 & 0.0294 & 2.802\textdegree & 5.663\textdegree \\
\addlinespace[2pt]
\hdashline
\addlinespace[2pt]
\textbf{12L-768D} & \textbf{0.0165} & \textbf{0.0287} & \textbf{2.299\textdegree} & \textbf{4.816\textdegree} \\
\bottomrule
\end{tabular}
\label{tab:ablation_dit}
\end{subtable}
\vspace{0.2cm}
\caption{\textbf{Ablation studies on Epic-Kitchens.} (a)~Scene encoder
comparison with a fixed 12L-768D DiT. (b)~DiT scaling with a fixed PTV3
encoder. Lower is better. L and D denote the number of layers and
embedding dimension of the DiT, respectively.}
\vspace{-0.2cm}
\label{tab:ablation}
\end{table*}

\subsection{Quantitative Results}
Table~\ref{tab:quantitative} reports all 6-DoF trajectory metrics on
Epic-Kitchens and HOT3D-Clips. On Epic-Kitchens, ObjectForesight-DiT
achieves the best translation and rotation accuracy overall, with an
ADE of 0.016\,m and an ARE of 2.30\textdegree. The autoregressive variant
underperforms the Constant Velocity baseline on most metrics, suggesting that without the diffusion
formulation the model struggles to capture the multimodal nature of
future object motion. In contrast, ObjectForesight-DiT surpasses
Constant Velocity on every metric, cutting ADE by 41\% and FDE by 45\%.
On HOT3D-Clips, ObjectForesight-DiT again leads across all metrics
(0.021\,m ADE, 8.92\textdegree\ ARE). The gap between Constant Velocity
and the learned methods is considerably larger on this dataset, indicating that HOT3D-Clips contains more
complex object motions that simple linear extrapolation cannot capture.
In the video-generation comparison on Epic-Kitchens, ObjectForesight-DiT
substantially outperforms Ray3 across all metrics (0.029\,m vs.\
0.084\,m ADE; 7.29\textdegree\ vs.\ 12.86\textdegree\ ARE),
reinforcing the benefit of predicting motion directly in SE(3) rather
than inferring it from synthesized frames. We present these results in a separate section of the table, as this evaluation is conducted on a subset of the data due to the high computational cost and manual effort required by the video generation model.

We additionally ablate the scene encoder backbone and the DiT model
scale (Table~\ref{tab:ablation}). For the encoder, we compare
\textit{PointTransformerV3} (PTV3), \textit{DGCNN},
\textit{PointNet++}, a simple sparse convolution network
(\textit{SparseConv}), and a variant with no scene encoding
(\textit{No-Encoder}). As shown in Table~\ref{tab:ablation_encoder},
PTV3 achieves the best results across all metrics. SparseConv performs comparably to or worse than
No-Encoder (e.g.\ 2.700\textdegree\ vs.\ 2.690\textdegree\ ARE),
indicating that a poorly suited geometric backbone can negate the
benefit of scene conditioning entirely, while a well-chosen encoder such
as PTV3 provides meaningful gains. Table~\ref{tab:ablation_dit} further
shows that scaling the DiT from 6L-384D to 12L-768D yields consistent
improvements across metrics, reducing ARE from 4.242\textdegree\ to 2.299\textdegree\
and ADE from 0.0193\,m to 0.0165\,m, clearly confirming that the task benefits
from increased model capacity.

\section{Conclusion}
We introduced the task of forecasting future 3D object motion directly
from passive human videos, framing it as an object-centric SE(3)
trajectory prediction problem. To support this task, we constructed a
large-scale dataset through automated segmentation, tracking, monocular
reconstruction, and pose alignment, yielding millions of metrically
grounded trajectories across diverse everyday manipulations without
requiring any manual annotation or motion capture.
\method\ combines monocular geometry, recent motion history, and local
scene structure within a diffusion-based transformer to produce
multimodal, physically consistent trajectory predictions. The diffusion
formulation is central to this capability, allowing the model to capture
the inherent uncertainty of future object motion rather than collapsing
to a single deterministic output. Experiments show that \method\
outperforms autoregressive and video-generation baselines across different metrics.

Our current formulation is limited to rigid objects and short prediction
horizons. Natural extensions include adopting more expressive
representations such as articulated kinematic models or learned
deformation fields, exploring longer prediction horizons, and
integrating with downstream planning or manipulation policies. More
broadly, our results establish a foundation for scalable, object-centric
3D dynamics modeling and point toward richer predictive models of
physical interaction.


%
%
\bibliographystyle{splncs04}
\bibliography{main}

\clearpage
\section{Appendix}
\setcounter{page}{1}
\appendix
\section*{Contents in the supplementary}

Please refer to our website \url{https://objectforesight.github.io/} for detailed qualitative results and videos. In the subsequent sections of this appendix, we elaborate on our dataset curation of 3D object poses from monocular egocentric videos, provide additional ablation studies that complement the results in the main paper, and expand on how we use scene context in our method. 

\section{Additional Details of the Data Curation Pipeline}

We elaborate on the stages of the data curation pipeline summarized in Sec. \ref{sec:data}, focusing on the heuristics, constraints, and cross-stage checks that improve data quality for pose trajectories.

\subsection{Presence Filtering and Initialization}
To mitigate false segmentations from EgoHOS, we aggregate interaction signals over each clip. We apply run-length smoothing to binary hand/object presence indicators using a threshold proportional to the clip length. This process fills brief detection gaps and eliminates false short-duration positives. The resulting smoothed signals serve two critical functions: they act as execution gates to ensure downstream modules run only when targets are reliably present, and they guide the SAM initialization towards temporally stable windows, reducing error propagation without introducing long-term drift.

\subsection{Robust 2D Tracking}

We augment the standard SAM2 tracking pipeline with a multi-stage regularization protocol that promotes temporal stability and suppresses duplicate object instances.

\noindent\textbf{Point Sampling Strategy.} To initialize and guide the model, we employ a robust sampling strategy. Positive points are sampled from the segmented object mask. To prevent mask leakage into the surrounding context, we explicitly sample negative points from three regions: detected hand masks, other object masks (if present), and a dilated background band surrounding the target object's mask.

\noindent\textbf{Temporal Stability and Consensus.} To mitigate per-frame segmentation noise, we construct a short-window consensus mask. When individual frame proposals are noisy, this consensus serves as a high-confidence positive prior. Furthermore, we apply mild morphological opening and closing to eliminate isolated speckles and smooth boundaries. 

\noindent\textbf{Trajectory Linking and De-duplication.} We associate object components across frames using greedy Intersection over Union (IoU) matching. To handle brief occlusions or detection failures, we permit a small gap tolerance in the temporal sequence. Tracks that fail to meet a minimum length requirement are discarded as noise. Simultaneously, we perform de-duplication within each video clip. If a new object proposal overlaps with an existing active track above a defined IoU threshold within a short temporal window, it is rejected. This ensures that the system maintains unique, distinct identifiers for each object instance.

\noindent\textbf{Initialization.} When multiple seeds are available, we prioritize candidates with the largest temporally stable area. Propagation is executed bidirectionally to maximize tracking duration.

\subsection{Quality Filtering and Selection}
We implement a two-stage filtering protocol to ensure only viable candidates reach the reconstruction stage.

\noindent\textbf{Manipulation Gate.} We employ a strict video-level gate using InternVL3 to filter out static or irrelevant objects. This module operates on object-highlighted visual summaries derived from the input track, rather than raw frames. Only tracks exhibiting active manipulation are retained.

\noindent\textbf{Clean-View Selection.} For the remaining valid tracks, we categorize frames into \textit{Partial/Invalid} (occluded, blurred, or insufficient resolution) and \textit{Clean} (unambiguous shape). Only clean frames are selected for geometry estimation. Input crops include a context margin to preserve local semantic cues.

\subsection{Reconstruction Preparation}

We prepare the data for 3D reconstruction through a sequence of filtering and completion steps.

\noindent\textbf{Frame Selection and Background Removal.} For the TRELLIS model, we select optimal "clean" frames based on foreground area size, excluding statistical outliers to maximize geometric consistency. Background clutter is masked out to isolate the object on a neutral canvas, enhancing texture and shape recovery.

\noindent\textbf{Amodal Mask Generation.} Separately, we use Diffusion-VAS to generate amodal masks. The segmentation masks contain holes or cutouts wherever the object is blocked by hands or other interactions. Diffusion-VAS corrects this by estimating the complete, physical shape of the object, filling in the missing regions. This ensures that we recover the full object silhouette, which is essential for accurate pose estimation and tracking in later steps of the pipeline.

\subsection{Pose Estimation and Tracking}
We adapt FoundationPose to recover robust 6DoF object trajectories, utilizing camera intrinsics, extrinsics, and dense depth maps provided by SpaTrackerV2. We add the following specific safeguards:

\noindent\textbf{Scale Estimation and Locking.} To handle monocular scale ambiguity, we lock the mesh diameter after the initial depth-to-mesh alignment. Subsequent residuals are normalized by this fixed diameter to ensure consistent error scoring across objects of varying sizes.

\noindent\textbf{Initialization Stress-Test.} To prevent tracking failures from the start, we do multi-view initialization of object pose. Each potential initial frame undergoes a brief ``refine-and-validate" optimization loop that jointly minimizes depth alignment error and maximizes silhouette consistency. Initial views yielding high depth alignment errors or silhouette inconsistencies are rejected.

\noindent\textbf{Bidirectional Tracking and Re-registration.} Tracking proceeds bidirectionally (forward and backward) from the optimal seed, with the estimator explicitly re-centered at the anchor frame before each pass. To detect and correct drift, we compute a suite of complementary consistency terms at every step:
\begin{itemize}
    \item \textbf{Silhouette Metrics:} We monitor Intersection over Union (IoU) with specific penalties for \textit{overflow} (mesh projection exceeding the mask) and \textit{underfill} (mesh projection failing to cover the mask).
    \item \textbf{Geometric Residuals:} We track the error between the rendered mesh depth and the observed sensor depth.
    \item \textbf{Motion Monitors:} We apply conservative thresholds on rotation and translation deltas to flag physically implausible jumps.
\end{itemize}
Re-registration is triggered by compounded evidence from these metrics, allowing the system to curb drift under heavy occlusions or rapid egocentric motion.

\section{Window-Size Ablation Studies}

In this section, we conduct a series of ablation studies to evaluate the contribution of different values of context length ($C$) and prediction horizon ($H$) and validate the design choices of our proposed framework.

\subsection{Ablation Study on Context Length}
\label{subsec:ablation_context}

\begin{table}[htbp]
\centering
\scriptsize
\setlength{\tabcolsep}{8pt} 
\begin{tabular}{ccccc}
\toprule
$C$ & ADE\,$\downarrow$ & FDE\,$\downarrow$ & ARE$\,\downarrow$ & FRE$\,\downarrow$ \\
\midrule
1 & 0.026 & 0.038 & 7.97\textdegree & 12.36\textdegree \\
2 & 0.021 & 0.033 & 7.61\textdegree & 12.14\textdegree \\
3 & \textbf{0.016} & \textbf{0.029} & \textbf{2.30\textdegree} & \textbf{4.82\textdegree} \\
5 & 0.025 & 0.035 & 7.69\textdegree & 11.68\textdegree \\
10 & 0.027 & 0.038 & 8.09\textdegree & 12.21\textdegree \\
\bottomrule
\end{tabular}
\vspace{0.1cm}
\caption{
\textbf{Ablation studies on the number of context frames.}
We evaluate the impact of context length $C$ on pose prediction accuracy with a fixed prediction horizon of $H=8$.
}
\vspace{-0.7cm}
\label{tab:P-abblations}
\end{table}

To determine the optimal temporal receptive field for our method, we conducted an ablation study on the number of input context frames $C$. We evaluated the model's performance by varying $C \in \{1, 2, 3, 5, 10\}$ while maintaining a fixed prediction horizon of $H=8$. To ensure a consistent evaluation benchmark across all configurations, the validation set was constructed using the maximum context length ($C=10$). For models trained with shorter contexts, we trimmed the input sequences accordingly, ensuring that all models predicted the exact same target frames based on the appropriate historical window. The results of this experiment are summarized in Table~\ref{tab:P-abblations}.

As illustrated in Table~\ref{tab:P-abblations}, we observe that increasing the context information initially improves prediction accuracy. The performance improves significantly as $C$ increases from 1 to 3, with $C=3$ achieving the lowest error rates across the majority of metrics, including an ADE of 0.018 and an ARE of 7.03\textdegree. This suggests that a context of three frames provides sufficient historical information to effectively capture the object's immediate trajectory and rotational dynamics.

However, increasing the context length beyond this point ($C=5$ and $C=10$) results in a performance degradation. For instance, at $C=10$, the ADE regresses to 0.027, and the ARE increases to 8.09\textdegree. We attribute this decline to two primary factors. First, longer context sequences are more susceptible to accumulated noise, which can distract the model from the most relevant recent motion cues. Second, an excessively long history may cause the model to overfit to past trajectories, hindering its ability to generalize to dynamic changes in pose movements or sudden shifts in direction. Consequently, we adopt $C=3$ as the default setting for our main method.

\subsection{Ablation Study on Prediction Horizon}
\label{subsec:ablation_horizon}

\begin{table*}[htbp]
\centering
\scriptsize
\setlength{\tabcolsep}{8pt}
\begin{tabular}{c cc cc cc cc}
\toprule
 & \multicolumn{2}{c}{Eval @ $H=4$} & \multicolumn{2}{c}{Eval @ $H=8$} & \multicolumn{2}{c}{Eval @ $H=16$} & \multicolumn{2}{c}{Eval @ $H=32$} \\
\cmidrule(lr){2-3} \cmidrule(lr){4-5} \cmidrule(lr){6-7} \cmidrule(lr){8-9}
Train $H$ & ADE & FDE & ADE & FDE & ADE & FDE & ADE & FDE \\
\midrule
4  & 0.016 & 0.023 & - & - & - & - & - & - \\
8  & \textbf{0.098} & \textbf{0.016} & \textbf{0.016} & \textbf{0.029} & - & - & - & - \\
16 & 0.015 & 0.020 & 0.022 & 0.034 & 0.034 & 0.055 & - & - \\
32 & 0.018 & 0.022 & 0.023 & 0.031 & \textbf{0.032} & \textbf{0.049} & \textbf{0.050} & \textbf{0.083} \\
\bottomrule
\end{tabular}
\vspace{0.1cm}
\caption{\textbf{Translation Error Analysis.} Comparison of ADE and FDE across models trained with different horizon lengths ($H$). Lower is better. Missing values (-) indicate the model cannot predict to that horizon.}
\vspace{-0.7cm}
\label{tab:trans_ablation}
\end{table*}

\begin{table*}[htbp]
\centering
\scriptsize
\setlength{\tabcolsep}{6pt}
\begin{tabular}{c cc cc cc cc}
\toprule
 & \multicolumn{2}{c}{Eval @ $H=4$} & \multicolumn{2}{c}{Eval @ $H=8$} & \multicolumn{2}{c}{Eval @ $H=16$} & \multicolumn{2}{c}{Eval @ $H=32$} \\
\cmidrule(lr){2-3} \cmidrule(lr){4-5} \cmidrule(lr){6-7} \cmidrule(lr){8-9}
Train $H$ & ARE\,$\downarrow$ & FRE\,$\downarrow$ & ARE\,$\downarrow$ & FRE\,$\downarrow$ & ARE\,$\downarrow$ & FRE\,$\downarrow$ & ARE\,$\downarrow$ & FRE\,$\downarrow$ \\
\midrule
4  & 1.51\textdegree & 2.39\textdegree & - & - & - & - & - & - \\
8  & \textbf{1.10\textdegree} & \textbf{1.95\textdegree} & \textbf{2.30\textdegree} & 4.82\textdegree & - & - & - & - \\
16 & 1.46\textdegree & 2.25\textdegree & 2.60\textdegree & 4.97\textdegree & 5.11\textdegree & 8.98\textdegree & - & - \\
32 & 1.87\textdegree & 2.40\textdegree & 2.77\textdegree & \textbf{4.68\textdegree} & \textbf{4.94\textdegree} & \textbf{8.46\textdegree} & \textbf{9.51\textdegree} & \textbf{15.67\textdegree} \\
\bottomrule
\end{tabular}
\vspace{0.1cm}
\caption{\textbf{Rotation Error Analysis.} Comparison of ARE and FRE across models trained with different horizon lengths ($H$). Lower is better.}
\vspace{-0.7cm}
\label{tab:rot_ablation}
\end{table*}

We further analyze the impact of the prediction horizon $H$ by training separate models with $H \in \{4, 8, 16, 32\}$ and a fixed input context length $C=3$. To ensure a fair comparison, the validation set is constructed using the maximum horizon ($H=32$); for models with shorter output capabilities, we crop the ground truth sequences to match their respective prediction lengths (4, 8, or 16 frames). This setup allows us to evaluate how training on different temporal lengths affects performance at various evaluation horizons.

Table~\ref{tab:trans_ablation} and Table~\ref{tab:rot_ablation} summarize the results for translation (ADE/FDE) and rotation (ARE/FRE) errors, respectively. Columns indicate the evaluation horizon used, while rows represent the model's training configuration.

The results highlight a clear trade-off between short-term precision and long-term capability. Interestingly, the model trained with $H=8$ outperforms the model trained with $H=4$ when evaluated at the shorter horizon of $H=4$ (e.g., ADE decreases from 0.0161 to 0.0098). This suggests that training on a slightly longer horizon encourages the network to learn more robust motion dynamics, acting as a form of regularization that benefits short-term accuracy.

However, blindly increasing the training horizon is not always beneficial. The model trained with $H=32$ exhibits significantly higher error rates at shorter horizons ($H=4, 8$) compared to the $H=8$ model. This degradation likely stems from the optimization difficulty; the loss function for $H=32$ is averaged over a long sequence where errors naturally accumulate, potentially diluting the gradients for earlier frames. Conversely, for long-term predictions ($H=16$ and $H=32$), the model explicitly trained on the larger horizon ($H=32$) yields the superior performance. This is expected, as models trained with shorter horizons optimize for immediate accuracy and lack the supervisory signal required to maintain trajectory consistency over extended periods. Without the long-term loss component, these models suffer from severe error accumulation (drift) when extrapolating beyond their training window. The $H=32$ model, by contrast, learns to model global temporal dependencies, effectively trading off some short-term precision for long-term stability.

\section{Scene Encoding Pipeline}
\label{sec:scene_encoder}

We now describe how a raw depth observation is converted into the global scene embedding that conditions the temporal pose predictor. The pipeline proceeds in four stages: point cloud construction, per-point feature design, context-aware pooling, and conditioning injection into the diffusion transformer.

\subsection{Point Cloud Generation and Preprocessing}

Given a depth map and the associated camera intrinsics, we back-project each valid depth pixel into a 3D point in the camera coordinate frame using the standard pinhole model. The resulting points are then transformed into anchor-frame camera coordinates using the camera extrinsics at the anchor frame~$t_a$. To obtain a fixed-size input, we first randomly subsample down to a coarse cap, then apply voxel-grid downsampling with a cell size of $0.005$\,m to ensure approximately uniform spatial coverage. If the number of surviving points falls below a target count of $N{=}4096$, we restore it by interpolating between randomly selected point pairs. The output is a point cloud $\mathbf{X} \in \mathbb{R}^{N \times 3}$ with metric anchor-frame coordinates, independent of the original depth resolution or scene density.

\subsection{Object-Centric Dual-Coordinate Features}

Rather than supplying only anchor-frame XYZ coordinates as input features, we augment each point with its position expressed in the object's local reference frame. Given the 6-DoF object pose at the anchor frame, $\mathbf{T}_\text{cam}^{\text{obj}} \in SE(3)$, we compute the inverse $\mathbf{T}_\text{obj}^{\text{cam}} = (\mathbf{T}_\text{cam}^{\text{obj}})^{-1}$ and transform each point to obtain its object-centric coordinates. The resulting 6D per-point feature vector is:
\begin{equation}
    \mathbf{f}_i = \left[\, x_i^{\text{cam}},\, y_i^{\text{cam}},\, z_i^{\text{cam}},\; x_i^{\text{obj}},\, y_i^{\text{obj}},\, z_i^{\text{obj}} \,\right],
\end{equation}
where the first three components are the anchor-camera coordinates and the latter three encode the same point's position relative to the object's center and orientation. This dual representation provides the backbone with an explicit geometric prior: the object-frame channel enables it to distinguish points on the object surface from those in the surrounding scene without having to learn this invariance from data alone. These 6D features, together with the anchor-frame coordinates used for spatial indexing, are passed to a PointTransformerV3 backbone, which produces a per-point feature $\mathbf{h}_i \in \mathbb{R}^{d}$ for each input point.

\subsection{Context Vector}
\label{subsec:context_vector}

To condition both the point cloud backbone and the downstream pooling on the object's recent motion history, we construct a context vector $\mathbf{ctx} \in \mathbb{R}^{64}$. For each frame $k$ in the conditioning sequence $1{:}t_a$, we concatenate the 9D pose token $\mathbf{p}_k$ with the normalized bounding box $\mathbf{B}_k$ to form a 13D descriptor $[\mathbf{p}_k, \mathbf{B}_k] \in \mathbb{R}^{13}$. A shared linear layer projects each descriptor into a 64-dimensional embedding. The resulting sequence is aggregated via anchor-query cross-attention: the anchor token (frame $t_a$) serves as the query and attends to all $t_a$ conditioning tokens, with sinusoidal positional encodings based on the relative temporal offset of each frame from the anchor. This produces a single 64D vector that summarizes the object's recent trajectory and visual extent. The context vector is supplied to PTv3's adaptive normalization layers (PDNorm) during backbone processing, and is reused in the pooling stage described next.

\subsection{Object-Aware Attentive Pooling}

After backbone processing, per-point features are linearly projected to the model's embedding dimension ($d{=}768$). We aggregate these into the global scene embedding $\mathbf{z}_{\mathrm{geom}}$ using an \emph{object-aware attentive pooling} mechanism that combines content-based and geometry-based cues:
\begin{itemize}
    \item \textbf{Content score.} A query vector $\mathbf{q}$ is obtained by projecting $\mathbf{ctx}$ into the embedding space. The content logit for point $i$ is the scaled dot product $s_i^{\text{cnt}} = \mathbf{h}_i^\top \mathbf{q}\, /\, \sqrt{d}$.
    \item \textbf{Distance score.} Each point's post-backbone coordinate is transformed into the object frame, and its distance from the object origin is computed. A small MLP with a learnable temperature $\tau$ maps this distance to a scalar bias: $s_i^{\text{dist}} = \text{MLP}\!\bigl(-\|\mathbf{x}_i^{\text{obj}}\|\, /\, \exp(\tau)\bigr)$.
    \item \textbf{Aggregation.} The final score $s_i = s_i^{\text{cnt}} + s_i^{\text{dist}}$ is normalized via softmax over all points within each sample, and the global feature is computed as the attention-weighted sum $\mathbf{z}_{\mathrm{geom}} = \sum_i w_i \, \mathbf{h}_i$.
\end{itemize}
By combining semantic relevance (content score) with spatial proximity to the object (distance score), this pooling directs the encoder's attention toward the object and its immediate surroundings while retaining access to the broader scene context.

\subsection{Scene Conditioning via AdaLN-Zero}

The scene embedding $\mathbf{z}_{\mathrm{geom}}$ is injected into the diffusion transformer via Adaptive Layer Normalization with zero initialization (AdaLN-Zero). Specifically, $\mathbf{z}_{\mathrm{geom}}$ is projected and passed through a small MLP, then fused with the diffusion timestep embedding via a learned combination MLP to produce a single conditioning vector $\mathbf{c}_{\text{comb}} \in \mathbb{R}^{768}$. Each transformer block contains a per-layer MLP that maps $\mathbf{c}_{\text{comb}}$ to six modulation parameters: a scale $\gamma$, shift $\beta$, and gate $\alpha$ for both the self-attention and feed-forward sub-layers. The layer normalization in each block carries no learnable affine parameters; modulation is instead applied as $\hat{\mathbf{h}} = (1 + \gamma)\,\text{LN}(\mathbf{h}) + \beta$, with the sub-layer output scaled by the gate $\alpha$. All gate parameters are zero-initialized so that each block begins as an identity function, promoting stable early training. This design ensures that the scene geometry conditions the denoising process deeply and uniformly at every layer, rather than through a single additive bias.


\end{document}